\documentclass{article}

    \PassOptionsToPackage{numbers, compress}{natbib}
    \usepackage[preprint]{neurips_data_2024}
\usepackage[utf8]{inputenc} % allow utf-8 input
\usepackage[T1]{fontenc}    % use 8-bit T1 fonts
\usepackage{hyperref}       % hyperlinks
\usepackage{url}            % simple URL typesetting
\usepackage{booktabs}       % professional-quality tables
\usepackage{amsfonts}       % blackboard math symbols
\usepackage{nicefrac}      % compact symbols for 1/2, etc.
\usepackage{microtype}      % microtypography
\usepackage{xcolor}         % colors

\usepackage{graphicx}
\usepackage{multirow}
\usepackage{pifont}
\usepackage{tikz}
\usepackage{amsmath}
\usepackage{amssymb}
\usepackage{bm}
\usepackage{tcolorbox}

\usepackage{duckuments}
\usepackage{subfiles}

\def\datasetName{APU }
\newcommand{\crossmark}{\ding{53}}
\newcommand{\halfmark}{\checkmark\kern-1.1ex\raisebox{.7ex}{\rotatebox[origin=c]{125}{--}}}
\newcommand{\norm}[1]{\left\lVert#1\right\rVert}

\title{Ask, Pose, Unite: Scaling Data Acquisition for Close Interactions with Vision Language Models}

\author{%
  Laura Bravo-S\'anchez \\
  Stanford University \\
  \texttt{lmbravo@stanford.edu} \\
  \And
  Jaewoo Heo\\
  Stanford University \\
  \texttt{jeffheo@stanford.edu} \\
  \And
  Zhenzhen Weng \\
  Stanford University \\
  \texttt{zzweng@stanford.edu} \\
  \And
  Kuan-Chieh Wang \\
  Snap Research \\
  \texttt{jwang23@snapchat.com} \\
  \And
  Serena Yeung-Levy \\
  Stanford University \\
  \texttt{syyeung@stanford.edu} \\
}

\begin{document}

\maketitle

\begin{figure}[ht]
    \centering
    \includegraphics[width=\textwidth]{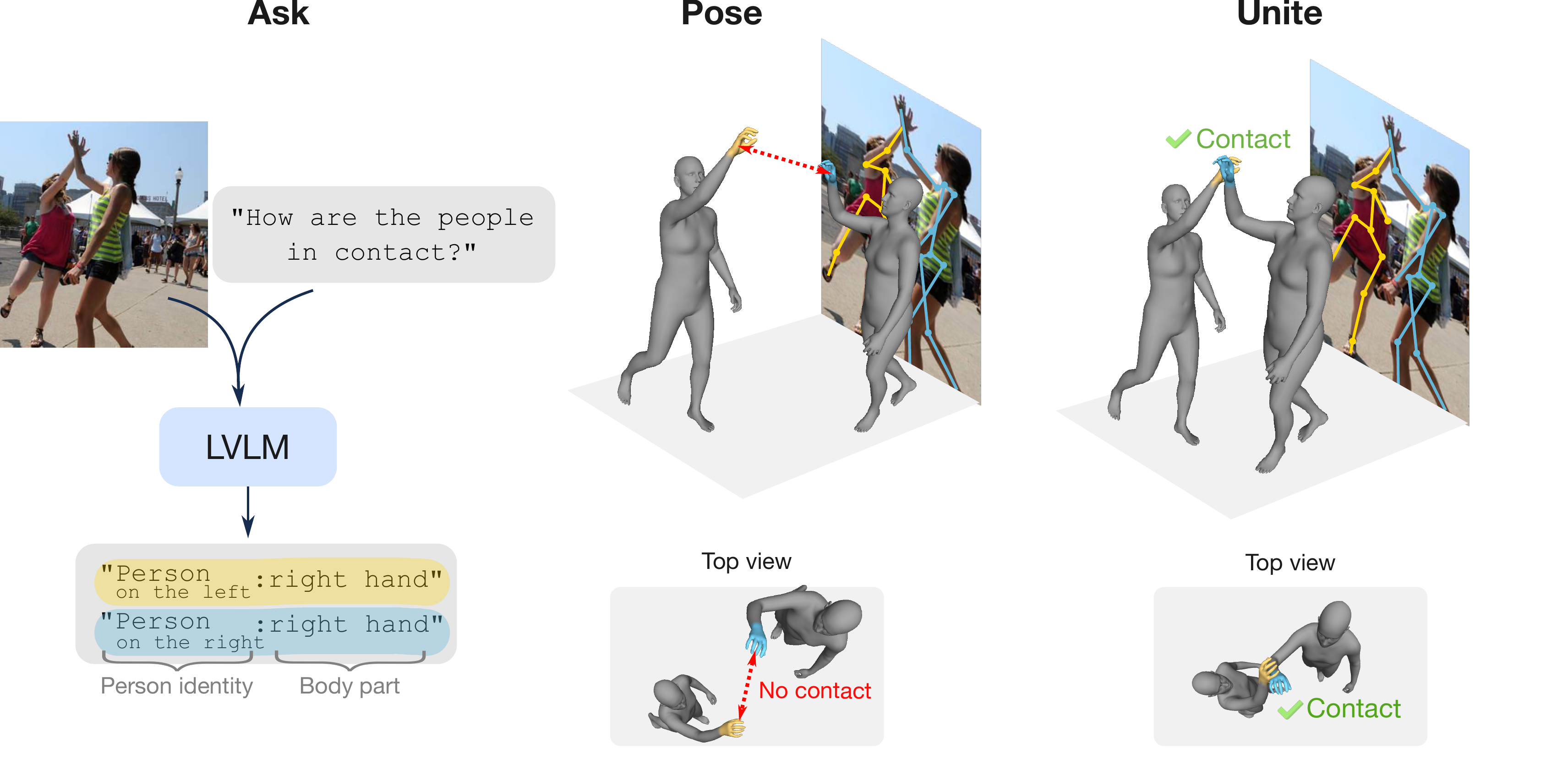}
    \caption{\textbf{Ask, Pose, Unite.} We scale data acquisition for close interactions by \textit{Ask}ing a Large Vision Language Model (LVLM) to identify contact points between people via language descriptions of the body parts that are touching. We \textit{Pose} 3D meshes in the scene with predicted 2D keypoints and \textit{Unite} the meshes in 3D by constraining an optimization of the mesh parameters with the predicted contacts. Through this data generation method we curate the Ask Pose Unite (APU) Human Mesh Estimation dataset for close interactions.}
    \label{fig:pull}
\end{figure}

\begin{abstract}
   Social dynamics in close human interactions pose significant challenges for Human Mesh Estimation (HME), particularly due to the complexity of physical contacts and the scarcity of training data. Addressing these challenges, we introduce a novel data generation method that utilizes Large Vision Language Models (LVLMs) to annotate contact maps which guide test-time optimization to produce paired image and pseudo-ground truth meshes. This methodology not only alleviates the annotation burden but also enables the assembly of a comprehensive dataset specifically tailored for close interactions in HME. Our Ask Pose Unite (APU) dataset, comprising over 6.2k human mesh pairs in contact covering diverse interaction types, is curated from images depicting naturalistic person-to-person scenes. We empirically show that using our dataset to train a diffusion-based contact prior, used as guidance during optimization, improves mesh estimation on unseen interactions. Our work addresses longstanding challenges of data scarcity for close interactions in HME enhancing the field’s capabilities of handling complex interaction scenarios. See our project website for the dataset, code, and contact prior: \url{https://laubravo.github.io/apu_website}.%Our dataset, code, and contact prior will be made publicly available.
\end{abstract}

\section{Introduction}
\label{sec:intro}

% Understanding human behavior is a fundamental pursuit indispensable for a myriad of fields such as socially aware robotics, patient-caregiver interactions in healthcare, parent-child interactions in psychology, etc. Central to this endeavor is the study of close interactions between individuals, which serve as a cornerstone for deciphering the intricacies of human dynamics. In recent years, the field of Human Mesh Estimation (HME) has emerged as a critical area of research aimed at deciphering complex scenes by harnessing the inherent priors encoded within parametric models. However, despite recent advancements in estimating multiple people in a scene, current HME methods face significant challenges when confronted with close interactions between individuals because the nuances of close contacts requires to address the problem in 3D, which is difficult because there's a scarcity of ground truth 3D meshes.

Understanding human behavior is fundamental for many fields, such as socially aware robotics, patient-caregiver interactions in healthcare, and parent-child interactions in psychology. Central to this pursuit is the study of close interactions between individuals, which are crucial for deciphering the complexities of human dynamics. The field of Human Mesh Estimation (HME) has emerged as a promising approach to study these dynamics, leveraging parametric models to interpret intricate scenes involving multiple people. However, despite significant advancements in multi-person HME, current methods struggle when faced with close interactions. This is because accurately reasoning about contacts requires a deep understanding of their 3D nature—how people touch and interact in three-dimensional space. Capturing this type of training data is particularly challenging due to the scarcity of ground truth 3D meshes, which are essential for precise estimation and analysis. The lack of detailed 3D data leaves gaps in the ability to model the subtleties of human contact, resulting in less accurate and reliable HME outcomes. Addressing this data scarcity is crucial for advancing the field and improving the handling of complex interaction scenarios.

Recent efforts \cite{Fieraru_2020_CVPR,yin2023hi4d,guo2022multi} have successfully acquired ground truth data for closely interacting scenes using motion capture (mocap) systems. Although effective, these systems are costly and limit the dataset's scope. Typically, these datasets feature only two subjects at a time, are confined to indoor lab environments, and cover a restricted set of predefined actions. Other approaches \cite{Fieraru_2020_CVPR,buddi} have proposed using weak supervision to avoid the need for mocap by formulating contact as a matching problem between surface body regions of the SMPL model \cite{loper2023smpl} in the form of binary contact matrices. While promising, this approach requires manual annotation of each contact region, limiting the scalability and integration into existing HME pipelines.

More recently, Müller et al. \cite{buddi} successfully generated pseudo-ground truth meshes from manually annotated image-contact matrix pairs from the FlickrCI3D dataset \cite{Fieraru_2020_CVPR}, which they used to train a diffusion-based contact prior for HME. The key insight for this approach was that the additional data enabled the contact prior to learn more meaningful contacts, significantly enriching the training set. However, despite these advancements, the contact prior still faces challenges with complex or out-of-distribution interaction scenarios. Problem compounded by the high cost of sourcing relevant images and manually annotating people interacting with their contact matrices. This highlights the ongoing need for solutions that can reflect the wide range of interactions found in in-the-wild scenes. Further progress in HME will be facilitated by datasets that accurately mirror natural interactions, capturing the full spectrum of human dynamics in diverse and unstructured environments.

% Paragraph for what we do in this work: 1) we address this problem by introducing a novel data generation method for automatically creating paired pseudo-gt meshes for closely interacting scenes using LVLMs. 2) We curate a dataset \datasetName with a large diversity of natural interaction types from images depicting people interacting in-the-wild. 3) We validate in a case study that our method and dataset can be used to improve the predictions in a new setting by retraining a contact prior for HME.

In this work, we introduce a novel dataset and data generation method to increase the diversity of posed meshes interacting closely. We develop an innovative approach \textit{Ask, Pose, Unite} that leverages Large Vision Language Models (LVLMs) to automatically create paired pseudo-ground truth meshes for scenes with closely interacting individuals (see Figure. \ref{fig:pull}). Our method enables the generation of accurate and detailed 3D representations of interactions without the need for costly and labor-intensive motion capture systems. To support this, we curate a comprehensive dataset, the Ask Pose Unite (\datasetName) dataset, which features a large diversity of natural interaction types. We source these types from in-the-wild images depicting people involved in close contact, capturing the complexity and variability of real-world human dynamics. By including a wide range of interaction scenarios, our dataset provides a robust foundation for training Human Mesh Estimation (HME) models for interacting humans. We validate the effectiveness of our data generation method and dataset through a case study, demonstrating that retraining a contact prior for HME with our data significantly improves prediction accuracy in new settings with uncommon actions. This validation shows that our approach not only enhances the quality of HME models but also ensures their adaptability to diverse interaction scenarios.

Our contributions can be summarized as follows: \textbf{(1)} We propose a novel data generation method for close interactions that leverages noisy automatic annotations to scale data acquisition, producing pseudo-ground truth meshes from in-the-wild images. \textbf{(2)} We curate \datasetName, a dataset of paired images and pseudo-ground truth meshes featuring a diverse array of close interaction types and subjects. \textbf{(3)} As an application, we demonstrate that our data significantly enriches the representation space of a close contact prior for HME, improving accuracy particularly for less common interaction scenarios in a case study of the NTU RGB+D 120 dataset.

\section{Related Work}
\label{sec:rw}

% \subsection{Generative models as priors and optimization-based multi-hme.}

\begin{figure}
    \centering
    \includegraphics[width=0.9\linewidth]{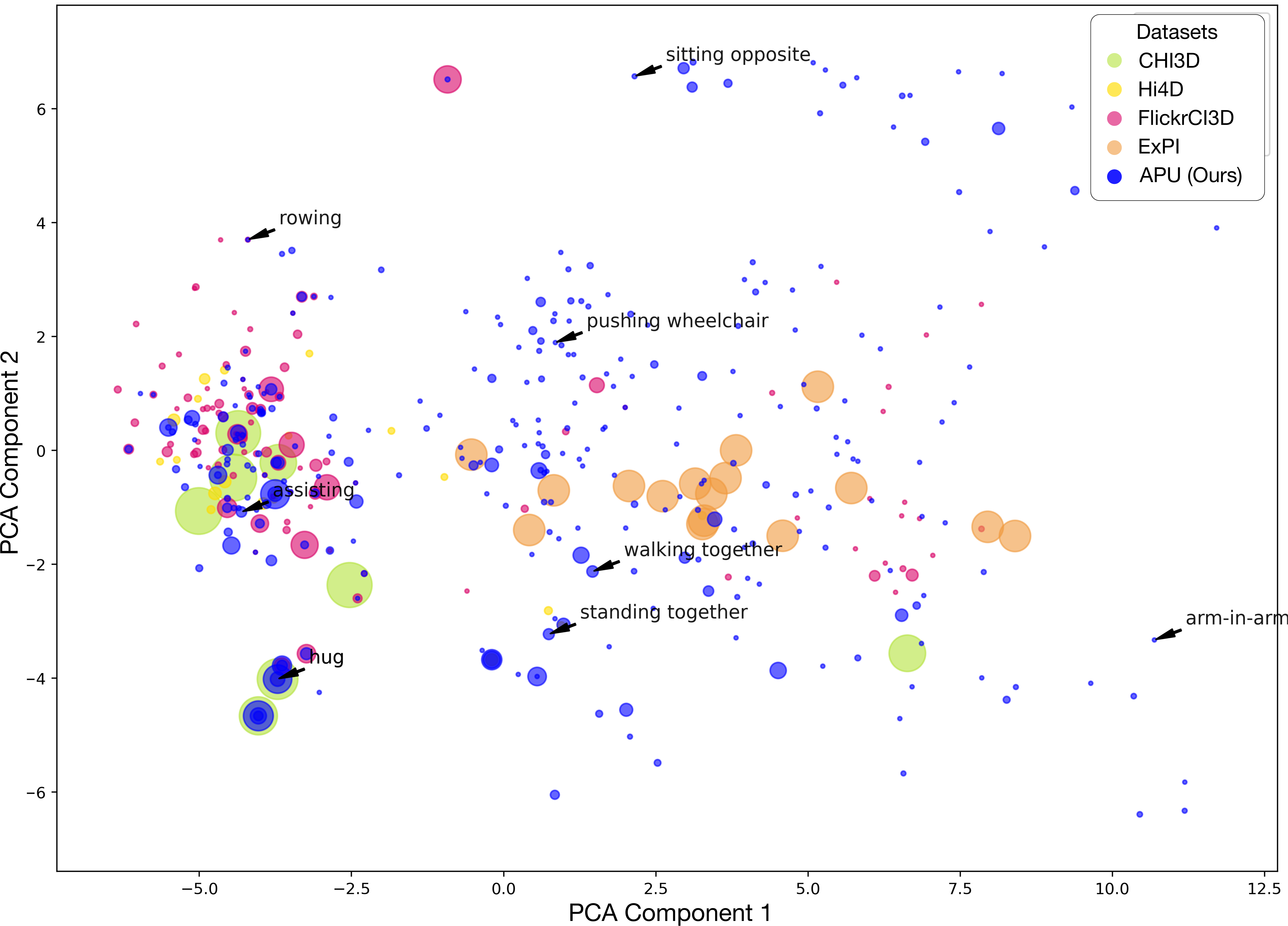}
    \caption{\textbf{Distribution of interaction types.} First two principal components of CLIP text embeddings on interaction names and grouped descriptions for existing datasets—CHI3D, Hi4D, FlickrCI3D, and ExPi— and our dataset. Size of points indicate quantity of examples. Our \datasetName dataset contributes a wide range of interactions compared to existing datasets, increasing the diversity of both examples and types of interactions captured.}
    \label{fig:stats}
\end{figure}

\textbf{Data Diversity in Multi-person HME.}
% Datasets for multi-person hme
% Monocular Human Mesh Estimation is an underspecified problem, particularly challenging due to the difficulty of capturing 3D ground truth, especially for multiple people. This challenge is addressed by leveraging alternate supervision strategies, such as weak annotations and synthetic data, to learn powerful priors for accurate 3D reconstruction.
Monocular Human Mesh Estimation is an underspecified problem, particularly challenging due to the difficulty of capturing paired 2D to 3D ground truth, especially for multiple people. To address this challenge, various datasets have adopted alternate supervision strategies (see Table \ref{tab:rw}). Some datasets, \cite{patel2021agora,black2023bedlam} use synthetic data to bypass the difficulties of capturing real-world 3D ground truth, providing a large number of images and subjects. Others \cite{zheng2021deepmulticap,mehta2018single} restrict their settings to lab environments, capturing high-quality 3D data but a trade-off on diversity. Since the work by \cite{bogo2016keep}, single person HME methods have relied on weak supervision to overcome the scarcity of paired 2D to 3D ground truth, using body part segmentations \cite{kanazawa2018end}, 2D keypoints \cite{kolotouros2019spin,weng2022domain}, and priors based on mocap data \cite{pavlakos2019expressive,kocabas2020vibe,wang2023refit,dposer}. Our work extends this line of research by introducing a data generation method for paired 2D to 3D pseudo-ground truth.

% Notable optimization- and regression-based methods. Mention how most works focus on obtaining accurate individual poses and shapes but they lack understanding about interactions between people. Include a paragraph on generative models as priors.
Multi-person HME methods often process individuals independently, which can yield accurate predictions for isolated figures but fails to correctly position them relative to one another in world space. Recent advancements have addressed these issues by using a unified spatial framework for entire images \cite{li2022cliff}, jointly modeling scene and camera dynamics \cite{yao2023w}, tracking people across time \cite{yuan2022glamr,rajasegaran2022tracking,TRACE,goel2023humans}, harnessing all available data \cite{cai2024smpler}, and managing occlusions \cite{choi2022learning,khirodkar2022occluded,zhu2024dpmesh}. In contrast, single-stage approaches \cite{ROMP,jiang2020coherent,BEV,multi-hmr2024,sun2024aios,wang2024multi}, which predict all subjects simultaneously, have demonstrated superior performance in terms of spatial accuracy and scale consistency. Despite their effectiveness, these methods depend heavily on extensive datasets, which are scarce for interactions involving close proximity. Our work aims to enrich the data available for these scenarios, potentially enhancing the effectiveness of existing models.

\begin{table}[t]
\centering
\caption{Multi-person Human Mesh Estimation datasets. \crossmark: absent, \checkmark present, \halfmark: has some examples.}
\label{tab:rw}
\resizebox{\textwidth}{!}{%
\begin{tabular}{lcccccccc}
\hline
Dataset  & Source & Type & Size & Subjects & Subjects & Actions & \begin{tabular}[c]{@{}c@{}}Close\\ Interactions\end{tabular} & \begin{tabular}[c]{@{}c@{}} Contact\\ anns.\end{tabular} \\ \hline \hline
AGORA \cite{patel2021agora} & synth. & images & 18k & all ages & 5 - 15 & - & \crossmark & \crossmark \\
BEDLAM \cite{black2023bedlam} & synth. & images & 380k & adults & 1 - 10 & -  & \crossmark & \crossmark \\
3DPW \cite{von2018recovering} & wild & video & 60 & adults &$\geq 2$ & - & \halfmark & \crossmark\\
MuPoTS-3D \cite{mehta2018single} & lab \& wild & video & 20 & adults & 3 & - & \halfmark & \crossmark \\
MultiHuman \cite{zheng2021deepmulticap} & lab & scans & 150 & adults & 1-3 & - & \halfmark & \crossmark \\
% JTA \cite{fabbri2018learning} &  & & & & & & &\\
\hline
ExPI\cite{guo2022multi} & lab & videos & 60k & adults & 2 & 16 & \checkmark & \crossmark \\
FlickrCI3D \cite{Fieraru_2020_CVPR} & wild & images & 10k & all ages & $\geq 2$ & - & \checkmark & \checkmark  \\
CHI3D \cite{Fieraru_2020_CVPR} & lab & video & 631 & adults & 2 & 8 & \checkmark & \checkmark \\
Hi4D \cite{yin2023hi4d} & lab & video & 100 & adults & 2 & 22 & \checkmark & \checkmark \\

\hline
\textbf{\datasetName (Ours)} & lab \& wild & images & 6209 & all ages & $\geq 2$ & >500 & \checkmark & \checkmark \\
\end{tabular}%
}
\end{table}

\textbf{Close interactions in HME.} Recently, studying close interactions in multi-person HME has become possible largely due to the introduction of new datasets (see Table \ref{tab:rw}). Lab-based datasets such as CHI3D \cite{Fieraru_2020_CVPR}, Hi4D \cite{yin2023hi4d}, and ExPI \cite{guo2022multi} provide 3D ground truth via capture systems with multiple calibrated cameras and propose methods to address heavy occlusions in close contact settings. However, this precision in annotations comes at the cost of the variety of scenes that can be captured, thus this works approach the large variability in interactions by recording a set of common actions. In this context, Fieraru et al. \cite{Fieraru_2020_CVPR} expand on the interaction types by tackling close interactions in-the-wild via weak supervision. In particular, by formulating proximity as a contact problem where the objective is to minimize the distance between surfaces in contact. \cite{Fieraru_2020_CVPR} introduce the FlickrCI3D dataset, a large collection of images from the internet where pairs of people in contact are manually annotated with contact maps—binary matrices that indicate which body parts are in contact. This approach has also been expanded to include self-contact \cite{muller2021self} and scene contact \cite{PROX,bhatnagar22behave,rich} scenarios. Despite the effectiveness of contact maps in lifting 2D information onto a 3D representation space, they are costly to annotate. By leveraging LVLMs, our work introduces an automatic method that sources contact maps directly from 2D images and extracts other relevant contextual information such as the type of interaction or descriptions of the scene. This automated approach reduces the annotation burden, scales up data acquisition, and enriches the dataset with varied interactions and valuable contextual details.

Learning from contact maps to model interactions has been more explored to capture human-object interactions, either by predicting contact maps directly from images \cite{chen2023detecting,tripathi2023deco} or by inferring object affordances from mesh estimates \cite{diller2023cg}. Concurrent to this work, \cite{subramanian2024pose} has very recently proposed a method to produce contact maps from LVLMs, but their approach is limited to the existing scope of close interaction datasets. In contrast, our dataset and method explicitly addresses the problem of data diversity by using LVLMs as part of a scalable data generation technique for HME in close interactions. Beyond human-object interactions, the focus has shifted towards understanding how individuals interact with their environments, such as improving motion realism through accurate ground-plane contact \cite{shin2023wham,tripathi2023deco,yang2023lemon}. Some works combat data scarcity through the use of synthetic data \cite{hassan2021populating,luo2022embodied}, pre-scanned scenes \cite{shen2023learning}, and by leveraging expert models in object detection and mesh reconstruction \cite{zhang2020perceiving}. Similarly, person-to-person interaction studies have also focused on predicting contact maps from images \cite{fieraru2021remips,cha20243d}, but they do not handle out-of-distribution or complex interactions well. In response, Müller et al. \cite{buddi} propose a diffusion-based contact prior trained with pseudo-ground truth 3D meshes created by constraining the optimization with manually annotated contact maps. Our work builds on this line of research by proposing a method to make a contact prior more robust to new interactions.

\textbf{LVLMs for 3D understanding.}
There is a growing line of research that employs LVLM's to obtain representations better aligned with real-world scenarios. Some works focus on 3D reasoning, utilizing LVLMs for the assessment of 3D reconstructions \cite{wu2024gpt}, motion generation \cite{qu2024gpt}, and enhancing the diversity of representation spaces \cite{han2023chorus,xie2024latte3d}. Others explore human-object contacts \cite{xu2024interdreamer,kim2024zero}, and directly reasoning about pose \cite{feng2024chatpose,tian2024hpecogvlm}. Our work contributes to this line of research by employing LVLMs as weak annotators to scale data generation and improve the modeling of close human interactions in 3D.
\section{Ask Pose Unite Dataset}
\label{sec:dataset}

\begin{figure}[t]
    \centering
    \includegraphics[width=\linewidth]{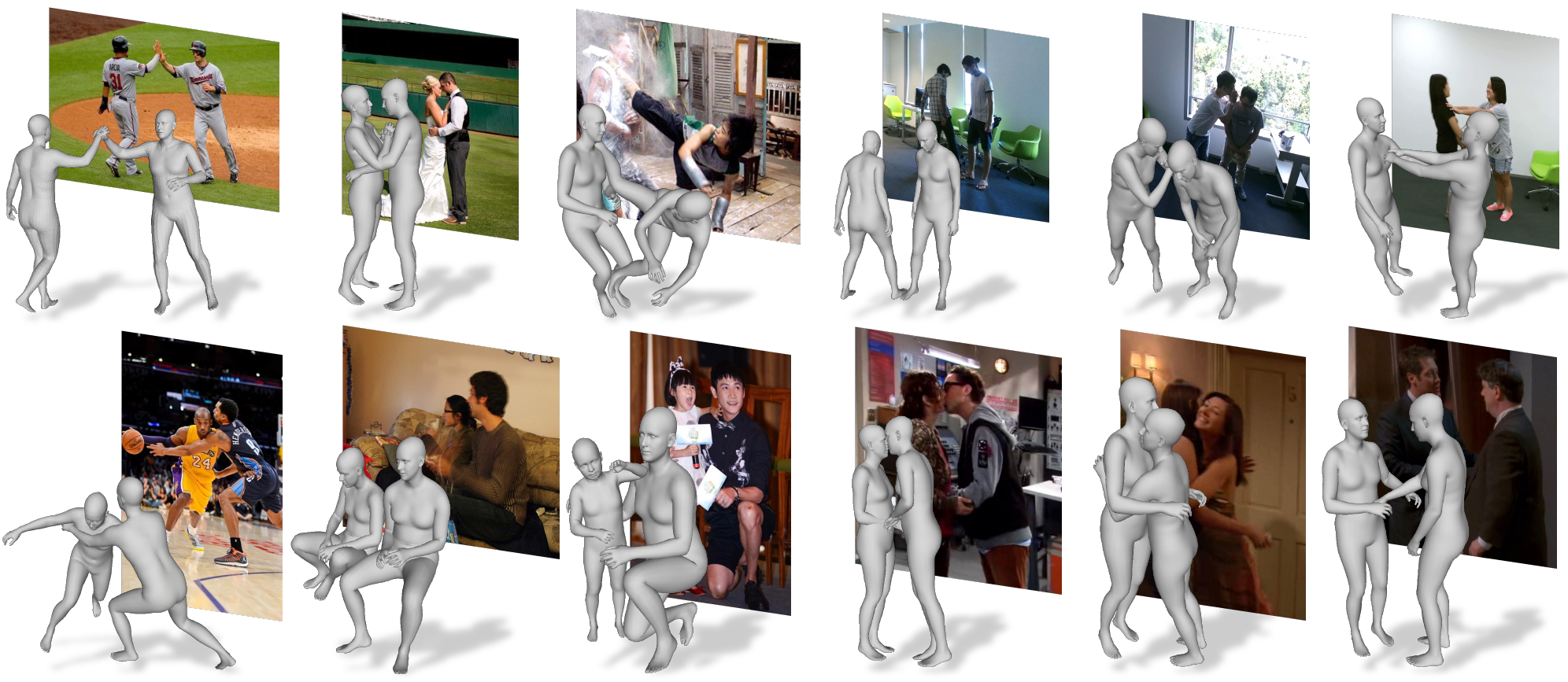}
    \caption{Examples of mesh pairs and images from our \datasetName dataset obtained with our data generation method. Note the variety of subjects, ages, interactions, and settings.}
    \label{fig:fits}
\end{figure}

% In this section we detail our data generation method and \datasetName dataset, a HME dataset for close person-to-person interactions with paired images and pseudo-ground truth meshes which we curate from in-the-wild images depicting closely interacting humans (see Figure \ref{fig:fits}).

\subsection{Dataset interaction type analysis}
We hypothesize that existing datasets that feature closely interacting humans often suffer from a lack of diversity and imbalance in their interaction types. We perform an analysis of the interactions in HME close interaction datasets using the representation space of CLIP \cite{clip} text embeddings as a proxy for analyzing the variety of interaction types across all datasets. For each dataset, including our \datasetName dataset, we curate a list of all unique interaction types as well as their respective frequencies. Then, we extract the CLIP text embeddings for all unique interaction types and visualize the principal components after PCA. Because FlickrCI3D lacks explicit classes we obtain per image descriptions with the BLIP-2 \cite{li2023blip} captioning model and group similar actions by  pattern matching on the action phrases. For \datasetName we use the interaction predicted by the LVLM. Figure \ref{fig:stats} shows our method's ability to increase data collection on interaction types that are typically under-represented in prior datasets. Our \datasetName dataset extends beyond the clusters formed by the other datasets, indicating that it includes novel interaction classes. We highlight some example interactions where the increase in diversity or points are noteworthy, such as "assisting", "hug", "arm-in-arm", among others. This straightforward experiment confirms our hypothesis that both our \datasetName dataset and data generation method are viable solutions for responding to the data scarcity problem in close interactions for HME.

\subsection{Dataset statistics}\label{subsec:sources}

We compile our \datasetName dataset by building on a key insights of prior works \cite{Fieraru_2020_CVPR,buddi}: using 2D images with weak labels to target interaction diversity. We have gathered more than 6,000 meshes paired with images, contact annotations, and natural language descriptions of the interactions from both laboratory and in-the-wild scenes, encompassing a variety of ages, subjects, and interactions (see Table \ref{tab:rw}). Figure \ref{fig:fits} shows examples of the images and mesh pairs from our dataset.

% \subsection{Data sources}\label{subsec:sources}
\begin{table}[t]
\centering
\caption{Data sources for our \datasetName dataset. Candidate pairs: possible people in contact from 2D keypoint distances. Final pairs: number of mesh pairs after automatic filtering.}
\label{tab:sources}
\resizebox{\textwidth}{!}{%
\begin{tabular}{lccccccc}
\hline
Data source & Images & Subjects & Actions & \# Actions & \# Subjects & Candidate pairs & Final pairs \\ \hline \hline
TV Interactions \cite{patron2012structured} & 8445 & adults & \checkmark & 4 & $\geq$ 2 & 5970 & 1679\\ 
Human Interaction Images \cite{tanisik2016facial} & 1177 & all ages & \checkmark & 7 & $\geq$ 2 & 954 & 282 \\
Relative Human \cite{sun2022putting} & 8740 & all ages & \crossmark & in-the-wild & $\geq$ 2 & 10577 & 2725 \\
NTU RGB+D 120 train \cite{liu2020ntu} & 3000 & adults & \checkmark & 11 & 2 & 2347 & 1523\\ 
% Unsplash \cite{unsplash_dataset} & 13813 & all ages & \crossmark & in-the-wild & $\geq$ 2 & 13813 &  \\
\end{tabular}%
}
\end{table}

To address the skewed distribution of interaction types in existing datasets, we curated the \datasetName dataset from two primary sources of images: those with and without action classes. Table \ref{tab:sources} shows the sources of data and their attributes, we introduce each data source below.

% existing image datasets, some of which include action classes, and Unsplash \cite{unsplash_dataset} a large-scale repository of stock images.

\textbf{TV Interactions} \cite{patron2012structured}. This dataset was collected from 300 video clips from 20 TV shows, containing 4 interactions: handshakes, hugs, high fives, and kisses, and clips without or with other interactions.

\textbf{Human Interaction Images} \cite{tanisik2016facial}. This dataset comprises images of facial expressions of people interacting. We selected 7 uncrowded action types: boxing-punching, handshaking, high-five, hugging, kicking, kissing, and talking.

\textbf{Relative Human} \cite{BEV}. This dataset focuses on multi-person scenes with people of all ages.

\textbf{NTU RGB+D 120 train} \cite{liu2020ntu}. A dataset for human action recognition with 3D joint annotations from Kinect sensors. We selected 11 of the 26 two-person interaction classes that involve close interactions and contact: punch/slap, kicking, pat on back, hugging, handshake, knock over, grab stuff, step on foot, high-five, whisper in ear, and support somebody. Then, we randomly selected a subset of 3000 images from frames where the subjects are in contact, determined by the 2D keypoints and 3D joint distance between people.

\subsection{Data generation method} \label{subsec:method}

\begin{figure}[t]
    \centering
    \includegraphics[width=\linewidth]{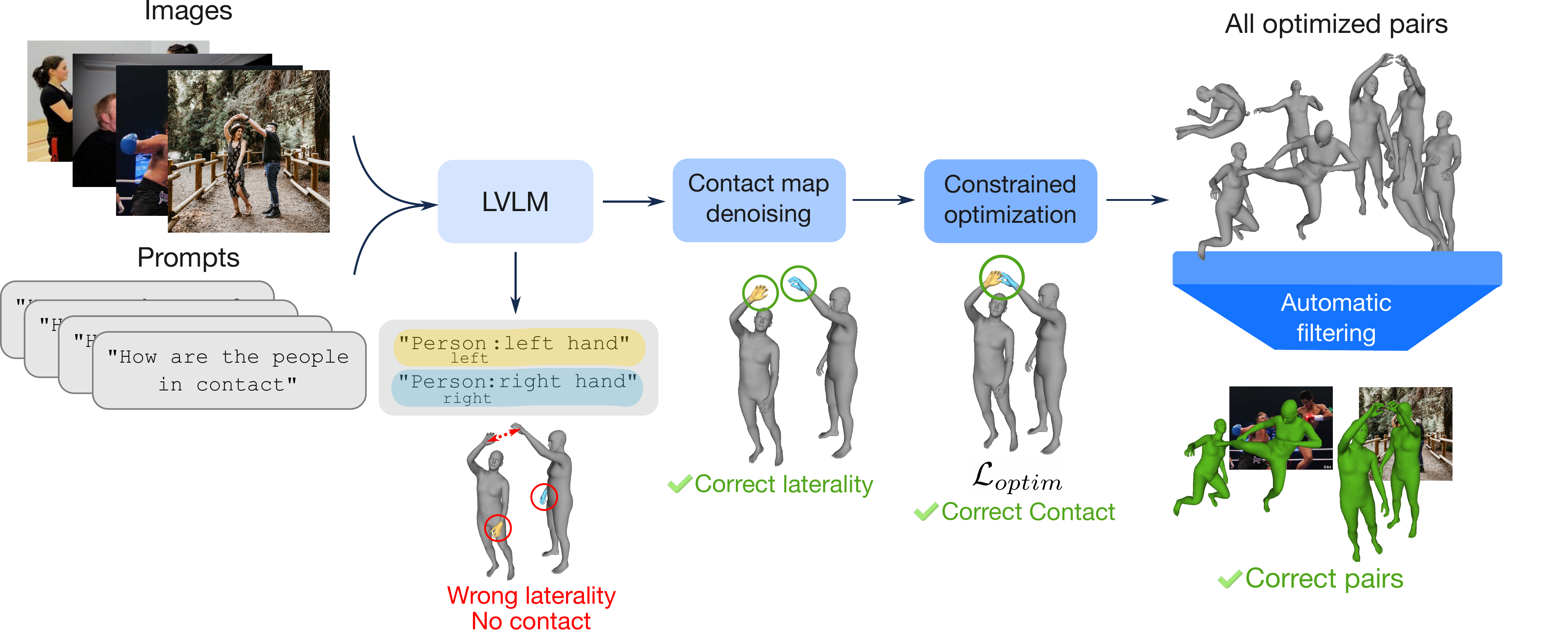}
    \caption{\textbf{Overview of our data generation method}. From any set of images we obtain pairs of people in contact and their pseudo-ground truth meshes. For candidate pairs of people in contact we query an LVLM for their contact maps, then denoise the laterality of the contact maps via predicted 2D keypoint chirality, we use the contacts to constrain the optimization of the mesh parameters and automatically filter out failure cases.}
    \label{fig:method}
\end{figure}

\textbf{Problem formulation.}
% HMR problem setup, close interaction setup for 2 people, + notation.
We aim to curate images depicting pairs of people closely interacting with well-reconstructed pseudo-ground truth meshes from any set of in-the-wild images. To achieve this, we propose a data generation method (Figure \ref{fig:method} outlines the main steps of our approach). Specifically, our goal is to locate pairs of closely interacting people within any set of images and produce mesh estimates for each pair. Since we only rely on weak supervision in the form of predicted contact maps, 2D keypoints, and interaction labels, we also aim to automatically select the well-reconstructed meshes.

In the context of a single image capturing a scene of close interactions between individuals, our objective is to fit a SMPL-X~\cite{pavlakos2019expressive} parametric 3D human mesh model for each individual $p$ to recover their pose $\theta_{p} \in \mathbb{R}^{21 \times 3}$ and shape $\beta_{p} \in \mathbb{R}^{10}$ parameters. We position each mesh in world coordinates by also estimating the root translation $\gamma_{p} \in \mathbb{R}^{3}$ and global body rotation $\phi_{p} \in \mathbb{R}^{3}$. Following previous work \cite{patel2021agora,BEV}, we support the prediction of multiple ages including children with the SMPL-XA model which adds an interpolation parameter $\sigma_{p}$ between the shape space of SMPL-X and SMIL~\cite{hesse2018learning}. In practice $\sigma_{p}$ is concatenated to the shape parameters such that $\beta_{p} \in \mathbb{R}^{11}$.

Given an unannotated image $I$ of two people closely interacting we aim to recover their meshes $M^a$ and $M^b$ by following an optimization of the parameters $\{\theta_{p}, \beta_{p}, \gamma_{p}, \phi_{p},\sigma_{p} \}_{p=a,b}$ under the constraint of a contact map $C$. Where $C \in \{0, 1\}^{R \times R}$ is defined as a guidance of which body surface regions are in contact. In particular, $C_{i, j} = 1$ indicates that region $r_i$ of $M^a$ is in contact with region $r_j$ of $M^b$.

\textbf{Candidate proposal.} For a set of images we obtain 2D keypoints and initial mesh predictions from from off-the shelf estimators. We propose as candidates all pairs of people with $k$ valid keypoints within a distance $d$ of each other and with mesh predictions aligned with the keypoints.

\textbf{LVLM contact map querying.} We employ a LVLM to automatically generate $C$. The inherent challenge of using LVLMs for this task lies in their low performance when grounding complex spatial relationships depicted in 2D images \cite{tong2024eyes}. Naively querying the LVLM often results in hallucinated or missing contacts leading to degenerate mesh predictions. We tackle this limitation by in-context prompting and denoising the contact maps (explained in the following section).

To query the LVLM for pairs of body parts that are touching in $I$ we first regroup the 75 body regions introduced in \cite{Fieraru_2020_CVPR} into coarser semantically meaningful sets. These sets correspond to the body parts: hand, arm, leg, thigh, chest, stomach, back, neck, face, head, foot, shoulder, elbow, knee, forearm, upper arm, and waist. In practice we re-map a body part $B_i$ to a list of corresponding body regions such that $B_i = \{r_1, r_2, ..., r_n\}$. When available, we ground the LVLM by incorporating a low-cost soft label $A$, which indicates the type of interaction depicted in the image. $A$ is a fundamental element of existing close interaction datasets and serves in this setting as a contextual prior.

\textbf{Contact map denoising.} We observe that even with contextual clues, LVLMs are unreliable when predicting the laterality of the body parts. We hypothesize that this problem arises from the model needing to reconcile two conflicting frames of reference: the visual perspective of the image and the anatomical orientation of the human body. For instance, a person's right hand may appear on the left side of an image. To correct these mistakes, we exploit the commonalities between estimated 2D pose keypoints and surface body regions. In particular, 2D keypoint estimation methods have been trained on larger sets of manual annotations that explicitly address the conflicting frames of reference.

Given two predicted body parts in contact with their body sides (either left, right, or both) $B_{i, side}$ and $B_{j,side}$, we use the normalized distance between the set of corresponding 2D keypoints for each body part to determine the chirality. Due to the difference in appearance between same-side and opposite-side contacts, we only compare the combinations that match the description. For example, if the prediction is $B_{i,left}$ and $B_{j,right}$, we evaluate two possible combinations: ($B_{i,left}$, $B_{j,right}$) and ($B_{i,right}$, $B_{j,left}$). We then select the combination with the closest normalized distance between the corresponding 2D keypoints.

\textbf{Constrained optimization.} Following prior work \cite{buddi,bogo2016keep}, we obtain pseudo-ground truth meshes $M^a$ and $M^b$ for a pair of people in contact using a two-stage optimization which takes as input estimated 2D keypoints, an initial estimate of the parameters $\{\tilde{\theta_{p}}, \tilde{\beta_{p}}, \tilde{\gamma_{p}} \}_{p=a,b}$, and $C$. In the first stage we optimize $\{\theta_{p}, \beta_{p}, \gamma_{p} \}_{p=a,b}$ given a contact loss $\mathcal{L}_{C}$ and other priors as guidance. We propose a soft version of the contact loss from \cite{buddi} to account for the uncertainty in the predicted contacts. $\mathcal{L}_{C} = \sum_{i,j} W_{ij} C_{ij} \min \limits_{v \in r_i, u \in r_j}{\norm{v - u}^2}$, where $u$ and $v$ are vertices, and $W_{ij} = $ is the normalized keypoint distance from the denoising step scaled by the LVLM's confidence for the contact.

As additional guidance for the optimization we use a pose prior based on a Gaussian Mixture Model $\mathcal{L}_{GMM}$ \cite{bogo2016keep}, an $L_2$ shape prior $\mathcal{L}_{\beta}$ that penalizes deviation from the SMPL-X mean shape, $\mathcal{L}_{\bar{\theta}}$ an $L_2$ loss that penalizes deviation from the initial pose $\tilde{\theta}_p$, and a 2D keypoint reprojection loss $\mathcal{L}_{J}$. In the second stage we fix $\beta_{p}$ and add $\mathcal{L}_{P}$ to resolve interpenetration between meshes \cite{buddi}.

The complete loss for the constrained optimization with values that re-weigh each term is: 
$\mathcal{L}_{optim} = \lambda_{J}\mathcal{L}_{J} + \lambda_{GMM} \mathcal{L}_{GMM} + \lambda_{\beta} \mathcal{L}_{\beta} +  \lambda_{P} \mathcal{L}_{P} + \lambda_{\bar{\theta}} \mathcal{L}_{\bar{\theta}}$

\textbf{Automatic filtering.} As a last step we implement a filtering strategy to remove incorrect mesh products from the optimization by thresholding the 2D keypoint reprojection loss of $M^a$ and $M^b$. We keep all instances with error less than 20 for both subjects.

\textbf{Implementation details.} We use GPT-4V \cite{achiam2023gpt} from 2024-04-09 as the LVLM and VitPose \cite{xu2022vitpose} as the keypoint estimator. For the constrained optimization with the generated contact maps we include Openpose \cite{openpose} as an additional keypoint estimator. We set $\lambda_C = 1.0, \lambda_J = 0.02$ for both stages and follow \cite{buddi} for all other hyperparameters. The optimization was processed on an internal Slurm Linux cluster with Nvidia A600, A100, and L40 GPUs. Due to mismatches between the assigned person identity between the LVLM and initial mesh predictions, for guidance during training we consider the minimum of $\mathcal{L}_C$ for both configurations of people.
\section{Case Study: Using \datasetName to improve estimation for novel interactions}
\label{sec:experiments}

The main advantage of our dataset and data acquisition method is to introduce training data from a larger variety of interactions for downstream HME models. We study the effect of enhancing the representation space one such model, a contact interaction prior from \cite{buddi}, with our dataset. Below we detail the contact prior model, training process and results.

% Overview: Introduce NTU RGB+D 120 as a proof of concept experiment of the performance boost that can happen when using our data and method for uncommon or new interactions. In this section we explain how we curate the test set of close interactions from the NTU RGB+D 120 dataset, how we train a diffusion-based contact prior with our dataset and the results across different interaction types.

\subsection{Close interaction NTU RGB+D 120 test set}
For every sequence in the dataset's original test set we label the contact frames with a combination of the distance between the annotated 3D joints, 2D keypoints from an off-the shelf estimator, and manual frame-level annotation. Then, we ensure the quality of the 3D joints both visually and by calculating the error between the 2D keypoints and reprojected joints. The final test set comprises 309 frames across 8 classes with a mean of 38.6 (SD: 16.3) frames per class.

%punch/slap, kicking, pat on back, hugging, handshake, knock over, grab stuff, step on foot, high-five, whisper in ear, and support somebody.

% Explain NTU and why selected it check overlap with dataset sources section. Explain that we had to curate it too because it can be noisy and how many images + classes with actual contact in test we ended up with.

\subsection{Enhancing a contact prior with the \datasetName dataset}
\paragraph{Diffusion model.}
% What is the diffusion model prior briefly, what are the losses to train.
As a contact prior we adapt BUDDI \cite{buddi}, a diffusion model conditioned on initial mesh estimates. During training the diffusion model gradually noises data samples to a point of randomness and then learns to reverse this process by denoising samples step-by-step until reaching a coherent structure. In particular, at each time step the noise level $t$ is uniformly sampled with $\epsilon_t \sim \mathcal{N}(0, \textbf{I})$ to obtain from a ground-truth sample $\textbf{x}_0$ the noisy sample $\textbf{x}_t = \sqrt{\sigma'_t} \textbf{x}_0 + \sqrt{1 - \sigma'_t} \epsilon_t$ with $\sigma'_t = \prod_{i=1}^{t} (1 - \sigma_t).$ BUDDI is trained to minimize $\mathbb{E}_{\textbf{x}_0 \sim p_{\text{data}}}    \mathbb{E}_{t\sim\mathcal{U}\{0, T\},\textbf{x}_t \sim q(\cdot | \textbf{x}_0)} ||BUDDI( \textbf{x}_t; t, \varnothing) - \textbf{x}_0 ||$. Specifically, an input sample $\textbf{x}_0$ corresponds to the input SMPL-XA parameters ${\phi_{p}, \theta_{p}, \beta_{p}, \gamma_{p}}$ for each person $p$. The loss to train the contact prior is $ \mathcal{L}_{prior} = \lambda_{\theta}\mathcal{L}_{\theta} + \lambda_{\beta}\mathcal{L}_{\beta} + \lambda_{\gamma}\mathcal{L}_{\gamma} + \lambda_{v2v}\mathcal{L}_{v2v} \text{, }$ where all terms are $L_2$ losses w.r.t the parameters and $\mathcal{L}_{v2v}$ is a squared $L_2$ loss on the vertices.

\paragraph{Inference with the contact prior.}
% Explain how inference with the contact prior works, what are the losses.
At test time, we perform a two-stage optimization process to obtain the mesh estimates $M^a$ and $M^b$ for the pair of people in an image similarly to section \ref{subsec:method}. However, we replace the contact map guidance with the trained contact prior. At each iteration we diffuse and denoise the current estimate $\textbf{x}_0$ with a noise level at $t=10$. The denoised estimate $\hat{\textbf{x}}_0$ regularizes the current estimate $\textbf{x}_0$ with an $L_2$ loss ${L_\text{diffusion}} = || \hat{\textbf{x}}_0  - \textbf{x}_0 || \text{.}$ In practice, the decoded parameters are penalized directly by $L_{\text{diffusion}} = \lambda_{\hat{\phi}} || \hat{\phi}_0 - \tilde{\phi} || + \lambda_{\hat{\theta}} || \hat{\theta}_0 - \tilde{\theta} || + \lambda_{\hat{\beta}} || \hat{\beta}_0 - \tilde{\beta} || + \lambda_{\hat{\gamma}} || \hat{\gamma}_0 - \tilde{\gamma} ||$. The contact prior offers enough guidance that the GMM pose prior is not needed. Thus, the complete loss function for the optimization is $\mathcal{L}_{optim} = \lambda_{J}\mathcal{L}_{J} + \lambda_{\tilde{\theta}} \mathcal{L}_{\tilde{\theta}} + \lambda_{P} \mathcal{L}_{P} + L_{\text{diffusion}}$. Where $\mathcal{L}_{\tilde{\theta}}$ is a prior to encourage the solution to be close to the denoised initialization.

\subsection{Results}
\textbf{Implementation details.}
% include total compute (gpu types and time used)
% specify hyper params and how they were chosen
% include data splits used for training for all the datasets
We follow the data preparation of \cite{buddi} and train the contact prior on the ground truth meshes of Hi4D and CHI3D, FlickrFits \cite{buddi} (the pseudo-ground truth derived from FlickrCI3D), and our \datasetName dataset. We train for 3k epochs on a batch composed of 40\% for FlickrFits and 20\% for the remaining datasets. We set the same hyperparameters as \cite{buddi} for training and inference with the contact prior and use an internal Slurm Linux cluster with Nvidia A600, A100, and L40 GPUs.

% explain baselines: BEV, contact maps, BUDDI
% explain metric Joint PA-MPJPE how it captures both people as a unit
\textbf{Baselines and metrics.} To validate the effect of training the contact prior with our \datasetName dataset, we compare the performance to several methods: BEV \cite{BEV}, a multi-human HME method that produces the initial mesh estimates input to the optimization; BUDDI \cite{buddi}, the state-of-the-art HME method for close interactions and (which is the contact prior trained without our dataset); and a baseline that uses contact matrices automatically generated from our method using the soft action labels (Auto CM). We evaluate errors between the predicted and ground-truth 3D joints using Mean Per Joint Position Error after aligning both people jointly with Procrustes Alignment (PA-MPJPE).

% Discuss tables \ref{tab:ntu}
\textbf{Discussion.} Table \ref{tab:ntu} shows the results for the close interaction categories of the NTU RGB+D 120 test set. The automatic contact map baseline (Auto CM) improves on most classes over the initial meshes from BEV. The contact prior benefits from the in-domain training data, performing better than BUDDI on all classes, and showing significant improvements on uncommon interactions such as step on foot, grab stuff, and support. Common actions, such as handshake and high-five, also benefit from a larger diversity of training examples. For qualitative results, see the supplementary material.

\begin{table}[t]
\centering
\caption{Joint PA-MPJPE results on close interaction NTU RGB+D 120 test set. PA-MPJPE: Joint two-person Procrustes aligned MPJPE. Auto CM: contact maps generated by our method. Best values in \textbf{bold}.}
\label{tab:ntu}
\resizebox{\textwidth}{!}{%
\begin{tabular}{lccccccccc}
\hline
Method & Overall & \begin{tabular}[c]{@{}c@{}}Pat on\\ back\end{tabular} & Handshake & \begin{tabular}[c]{@{}c@{}}Knock\\ over\end{tabular} & \begin{tabular}[c]{@{}c@{}}Grab\\ stuff\end{tabular} & \begin{tabular}[c]{@{}c@{}}Step on\\ foot\end{tabular} & High-five & Whisper & Support \\ \hline \hline
BEV & 110.9 & 93.5 & 117.8 & 113.0 & 109.0 & 107.7 & 108.8 & 129.9 & 115.7 \\
% v1
Ours (Auto CM) & 101.3 &  100.3 & \textbf{100.6} & 96.2 & 94.8 & 95.2 & 103.4 & 106.2 & 109.5  \\
BUDDI \cite{buddi} & 97.8 & 89.7 & 112.7 & 89.3 & 96.8 & 100.4 & 105.6 & 96.5 & 98.6 \\
% e563
Ours (Contact prior) & \textbf{91.9} & \textbf{86.7} & 101.2 & \textbf{87.8} & \textbf{89.7} & \textbf{90.9} & \textbf{94.1} & \textbf{94.6} & \textbf{94.8} \\

\end{tabular}%
}
\end{table}

\section{Conclusion}
\label{sec:conclusion}

In this paper, we address a key challenge in HME: data scarcity for new domains. We introduce a novel data generation method for close interactions, leveraging automatic annotations to produce pseudo-ground truth meshes from in-the-wild images. We curated \datasetName, a diverse dataset of paired images and pseudo-ground truth meshes, covering a wide range of close interaction types. We demostrate that our data can be used to improve HME methods for close interactions, particularly for less common interaction scenarios, on a case study of the NTU RGB+D 120 dataset.

\textbf{Limitations \& Ethical Concerns.}
Close interactions in HME is an ongoing line of research. Our automatic data generation method filters out many close interaction images even if they appear suitable, yet if the 2D keypoints, initial mesh estimation, or automatic contact maps are not all accurate, the images can be excluded. As improvements in the models that produce each of these components are made, the diversity of interactions will increase.

Due to budget constraints, we query a single LVLM, GPT-4V \cite{achiam2023gpt}, once per image to generate the contacts for our dataset. This approach may replicate the existing biases of this model in our dataset. Producing multiple outputs per image and querying multiple LVLMs could provide a measure of uncertainty to the predicted contacts, which could be integrated into our soft contact maps for improved robustness. We do not foresee significant risks of security threats or human rights violations in our work. However, the advancements in close interactions HME could be misused for creating misleading visual content, leading to potential harm or deception.
\section{Acknowledgements}
This work was supported by: the Fulbright U.S. Student Program sponsored by the U.S. Department of State and Fulbright Colombia (L. Bravo-S\'anchez).
Stanford Wu Tsai Human Performance Alliance (KC. Wang), and the National Science Foundation under Grant No. 2026498. Portions of the research in this paper used the NTU RGB+D (or NTU RGB+D 120) Action Recognition Dataset made available by the ROSE Lab at the Nanyang Technological University, Singapore. The authors thank the BUDDI authors for their great work and availability in answering questions. We also thank James Burgess for his invaluable discussions and comments on the draft.

\bibliographystyle{splncs04}
\bibliography{main}

\newpage
\clearpage
% Define the command for supplementary titles
\setcounter{page}{1}
\setcounter{section}{0}
 \vspace*{1.5cm} % Adjust vertical space before title
    \begin{center}%
      {\Large\bfseries Supplementary Material for Ask, Pose, Unite: Scaling Data Acquisition for Close Interactions with Vision Language Models \par}
    \end{center}%
    \par
    \vskip 1.5em
\renewcommand{\thefigure}{S\arabic{figure}} % LBS: added for clarity
\renewcommand{\thesection}{\Alph{section}}% This will prepend 'A' to the section number
\renewcommand\theHsection{\Alph{section}}

% The \author macro works with any number of authors. There are two commands
% used to separate the names and addresses of multiple authors: \And and \AND.
%
% Using \And between authors leaves it to LaTeX to determine where to break the
% lines. Using \AND forces a line break at that point. So, if LaTeX puts 3 of 4
% authors names on the first line, and the last on the second line, try using
% \AND instead of \And before the third author name.

% \author{%
%   Laura Bravo-S\'anchez \\
%   Stanford University \\
%   \texttt{lmbravo@stanford.edu} \\
%   % examples of more authors
%   \And
%   Jaewoo Heo\\
%   Stanford University \\
%   \texttt{jeffheo@stanford.edu} \\
%   \And
%   Zhenzhen Weng \\
%   Stanford University \\
%   \texttt{zzweng@stanford.edu} \\
%   \And
%   Kuan-Chieh Wang \\
%   Snap Research \\
%   \texttt{jwang23@snapchat.com} \\
%   \And
%   Serena Yeung-Levy \\
%   Stanford University \\
%   \texttt{syyeung@stanford.edu} \\
% }

% \thanks{} for astericks above name

\section{Overview}
This supplementary material provides detailed information about the \datasetName dataset, which is derived from our novel data generation method. Unlike traditional datasets compiled through direct data collection or benchmarking, our dataset uses automatic annotations and pseudo-ground truth meshes posed from in-the-wild images. This document includes the dataset datasheet, prompting strategy and additional analysis, highlighting the unique aspects and methodology behind our work. We have included a code.zip file with our GitHub repository containing the code and instructions for accessing the APU dataset.

\section{Dataset Datasheet}
%%%%%%%%%%%%%%%%%%%%%%%%%%%%%%%%%%%%%%%%%%
\subsection{Motivation}
\textbf{For what purpose was the dataset created?}
%{Was there a specific task in mind? Was there a specific gap that needed to be filled? Please provide a description.}
We created the APU dataset and our data generation method to diversify the paired image and mesh available for closely interacting humans.

\textbf{Who created this dataset}
% (e.g., which team, research group) and on behalf of which entity (e.g., company, institution, organization)?}
Members of the MARVL group at Stanford University. KC. Wang while at Stanford University, now at SNAP Research.

\textbf{Who funded the creation of the dataset?}
% {If there is an associated grant, please provide the name of the grantor and the grant name and number.}
 This work was supported by: the Fulbright U.S. Student Program sponsored by the U.S. Department of State and Fulbright Colombia (L. Bravo-S\'anchez); Stanford Wu Tsai Human Performance Alliance (KC. Wang), and partially by the National Science Foundation under Grant No. 2026498.
%%%%%%%%%%%%%%%%%%%%%%%%%%%%%%%%%%%%%%%%%%
\subsection{Composition}

\textbf{What do the instances that comprise the dataset represent (e.g., documents, photos, people, countries)?}
% { Are there multiple types of instances (e.g., movies, users, and ratings; people and interactions between them; nodes and edges)? Please provide a description.}
The basic data element is an image of a pair of people. This image can be complete or a portion of a larger image. For each pair of people we provide their posed meshes in SMPL-XA format, their keypoints and bounding boxes predicted by VitPose and Openpose, and the LVLM's output which includes the interaction type, description of the people and list of body parts in contact.

\textbf{How many instances are there in total?}
6209 instances of pairs of people interacting sourced from in-the-wild and laboratory images.

\textbf{Does the dataset contain all possible instances?}
%or is it a sample (not necessarily random) of instances from a larger set?}{ If the dataset is a sample, then what is the larger set? Is the sample representative of the larger set (e.g., geographic coverage)? If so, please describe how this representativeness was validated/verified. If it is not representative of the larger set, please describe why not (e.g., to cover a more diverse range of instances, because instances were withheld or unavailable).}
We source the images for the dataset from 4 existing datasets: TV interactions, Human interaction images, Relative human, and close interaction classes from NTU RGB+D 120. Each image may contain from 0 to multiple pairs of people interacting, we provide the instances with reconstructions that have a keypoint reprojection error less than 20.0 (see Table 2 of the main paper for more details).

For the NTU RGB+D 120 train set we randomly selected 3000 initial from a complete set of 3 frames per sequence that could contain people in contact. For the test set where we included a keyframe from each sequence where the subjects were in contact. We manually inspected all images from the final test set only.

\textbf{What data does each instance consist of?}
% “Raw” data (e.g., unprocessed text or images) or features?}{In either case, please provide a description.}
The raw data are the images from each data source. 

\textbf{Is there a label or target associated with each instance?}
%{If so, please provide a description.}
For each pair of people we provide the reconstructed meshes from our method.

\textbf{Is any information missing from individual instances?}
% {If so, please provide a description, explaining why this information is missing (e.g., because it was unavailable). This does not include intentionally removed information, but might include, e.g., redacted text.}
No

\textbf{Are relationships between individual instances made explicit (e.g., users’ movie ratings, social network links)?}
% {If so, please describe how these relationships are made explicit.}
We provide a data file with a dictionary for every interacting pair with: the name and bounding box of the original image, interaction type, 2D keypoints, SMPL-XA mesh parameters, and LVLM's output.

\textbf{Are there recommended data splits (e.g., training, development/validation, testing)?}
% {If so, please provide a description of these splits, explaining the rationale behind them.} 
We use all pairs for training except those from the NTU RGB+D test set.

\textbf{Are there any errors, sources of noise, or redundancies in the dataset?}
% {If so, please provide a description.}
As we generate pseudo-ground truth meshes these and the contacts and the keypoints may not correspond exactly to what is depicted in the image. We noticed that some images from Relative Human are duplicated under different names.

\textbf{Is the dataset self-contained, or does it link to or otherwise rely on external resources (e.g., websites, tweets, other datasets)?}
% {If it links to or relies on external resources, a) are there guarantees that they will exist, and remain constant, over time; b) are there official archival versions of the complete dataset (i.e., including the external resources as they existed at the time the dataset was created); c) are there any restrictions (e.g., licenses, fees) associated with any of the external resources that might apply to a future user? Please provide descriptions of all external resources and any restrictions associated with them, as well as links or other access points, as appropriate.}
We provide the links to download the original images from the source datasets. All other products are self-contained.

\textbf{Does the dataset contain data that might be considered confidential (e.g., data that is protected by legal privilege or by doctor-patient confidentiality, data that includes the content of individuals non-public communications)?}
% {If so, please provide a description.}
The images depict people and their faces which makes the data identifiable but all are within the public domain.

\textbf{Does the dataset contain data that, if viewed directly, might be offensive, insulting, threatening, or might otherwise cause anxiety?} No
% {If so, please describe why.}

\textbf{Does the dataset relate to people?}
% {If not, you may skip the remaining questions in this section.}
Yes

\textbf{Does the dataset identify any subpopulations (e.g., by age, gender)?}
% {If so, please describe how these subpopulations are identified and provide a description of their respective distributions within the dataset.}
No. We only specify that it contains all ages instead of only adults.

\textbf{Is it possible to identify individuals (i.e., one or more natural persons), either directly or indirectly (i.e., in combination with other data) from the dataset?}
% {If so, please describe how.}
The individuals can be identified by their faces.

\textbf{Does the dataset contain data that might be considered sensitive in any way (e.g., data that reveals racial or ethnic origins, sexual orientations, religious beliefs, political opinions or union memberships, or locations; financial or health data; biometric or genetic data; forms of government identification, such as social security numbers; criminal history)?}
% {If so, please provide a description.}
No
%%%%%%%%%%%%%%%%%%%%%%%%%%%%%%%%%%%%%%%%%%
\vspace{-5mm}
\subsection{Collection Process}

\textbf{How was the data associated with each instance acquired?}
% {Was the data directly observable (e.g., raw text, movie ratings), reported by subjects (e.g., survey responses), or indirectly inferred/derived from other data (e.g., part-of-speech tags, model-based guesses for age or language)? If data was reported by subjects or indirectly inferred/derived from other data, was the data validated/verified? If so, please describe how.}
The source images were directly observable and the products of our dataset were derived from our data generation method. We validated the quality of the posed meshes with a threshold on the 2D keypoint reprojection error. For the NTU RGB+D 120 test set we manually verified the quality of the 3D ground truth joints. 

\textbf{What mechanisms or procedures were used to collect the data (e.g., hardware apparatus or sensor, manual human curation, software program, software API)?}
% {How were these mechanisms or procedures validated?}
The source images were collected from existing image datasets and the products of our dataset were derived from our data generation method.

\textbf{If the dataset is a sample from a larger set, what was the sampling strategy (e.g., deterministic, probabilistic with specific sampling probabilities)?}
The pairs of interacting people are a subsample of all existing pairs in the images. Our strategy was to use our data generation method to select the pairs in contact with valid posed meshes.

\textbf{Who was involved in the data collection process (e.g., students, crowdworkers, contractors) and how were they compensated (e.g., how much were crowdworkers paid)?} Only the researchers were involved in the collection process.

\textbf{Over what timeframe was the data collected? Does this timeframe match the creation timeframe of the data associated with the instances (e.g., recent crawl of old news articles)?}
% {If not, please describe the timeframe in which the data associated with the instances was created.}
This dataset was collected in 2024, the original images correspond to publications from 2012 (TV Interactions), 2016 (Human Interaction Images), 2016/2019 (NTU RGB+D 120), and 2019/2022 (Relative Human). 

\textbf{Were any ethical review processes conducted (e.g., by an institutional review board)?}
% {If so, please provide a description of these review processes, including the outcomes, as well as a link or other access point to any supporting documentation.}
No

\textbf{Does the dataset relate to people?} Yes
% {If not, you may skip the remaining questions in this section.}

\textbf{Did you collect the data from the individuals in question directly, or obtain it via third parties or other sources (e.g., websites)?} We obtained the images from existing datasets

\textbf{Were the individuals in question notified about the data collection?}
% {If so, please describe (or show with screenshots or other information) how notice was provided, and provide a link or other access point to, or otherwise reproduce, the exact language of the notification itself.}
The images were collected from consenting individuals (NTU RGB+D 120) and public domain images.

\textbf{Did the individuals in question consent to the collection and use of their data?}
% {If so, please describe (or show with screenshots or other information) how consent was requested and provided, and provide a link or other access point to, or otherwise reproduce, the exact language to which the individuals consented.}
The images were collected from consenting individuals (NTU RGB+D 120) and public domain images.

\textbf{If consent was obtained, were the consenting individuals provided with a mechanism to revoke their consent in the future or for certain uses?}
% {If so, please provide a description, as well as a link or other access point to the mechanism (if appropriate).}
No, but access to the images can be removed. 

\textbf{Has an analysis of the potential impact of the dataset and its use on data subjects (e.g., a data protection impact analysis) been conducted?}
% {If so, please provide a description of this analysis, including the outcomes, as well as a link or other access point to any supporting documentation.}
No

%%%%%%%%%%%%%%%%%%%%%%%%%%%%%%%%%%%%%%%%%%
\subsection{Preprocessing/cleaning/labeling}

\textbf{Was any preprocessing/cleaning/labeling of the data done (e.g., discretization or bucketing, tokenization, part-of-speech tagging, SIFT feature extraction, removal of instances, processing of missing values)?}
% {If so, please provide a description. If not, you may skip the remainder of the questions in this section.}

\textbf{Was the “raw” data saved in addition to the preprocessed/cleaned/labeled data (e.g., to support unanticipated future uses)?}
% {If so, please provide a link or other access point to the “raw” data.}
The original images can be accessed throught the links we provide.

\textbf{Is the software used to preprocess/clean/label the instances available?}
% {If so, please provide a link or other access point.}
Yes, we will make our code available.

%%%%%%%%%%%%%%%%%%%%%%%%%%%%%%%%%%%%%%%%%%
\subsection{Uses}

\textbf{Has the dataset been used for any tasks already?}
% {If so, please provide a description.}
In the paper we show how the data can be used to train a contact prior for Human Mesh Estimation.

\textbf{Is there a repository that links to any or all papers or systems that use the dataset?}
% {If so, please provide a link or other access point.}
No. All works that use our dataset should cite us and this should be accessible through google scholar.

\textbf{What (other) tasks could the dataset be used for?}
We used the data from a 3D application, but it can also be used for 2D tasks like image generation and general 2D understanding of person-to-person interactions.

\textbf{Is there anything about the composition of the dataset or the way it was collected and preprocessed/cleaned/labeled that might impact future uses?}
% {For example, is there anything that a future user might need to know to avoid uses that could result in unfair treatment of individuals or groups (e.g., stereotyping, quality of service issues) or other undesirable harms (e.g., financial harms, legal risks) If so, please provide a description. Is there anything a future user could do to mitigate these undesirable harms?}
The dataset was not compiled to have an equal ethnic or demographic distribution, as such, downstream tasks should be aware of the possible sampling biases in the data.

\textbf{Are there tasks for which the dataset should not be used?}
% {If so, please provide a description.}
The dataset focuses on broadly on human interactions. It should not be used to generate any explicit or harmful content from the subjects in the images or any other subjects.
%%%%%%%%%%%%%%%%%%%%%%%%%%%%%%%%%%%%%%%%%%
\subsection{Distribution}

\textbf{Will the dataset be distributed to third parties outside of the entity (e.g., company, institution, organization) on behalf of which the dataset was created?}
% {If so, please provide a description.}
Any third party that applies for the data can use it for non commercial uses in accordance with the license.

\textbf{How will the dataset will be distributed (e.g., tarball on website, API, GitHub)}
% {Does the dataset have a digital object identifier (DOI)?}
Through the website \url{https://laubravo.github.io/apu_website} and Github repository. Which will link to the code and  detail the process for accessing the data, including an application form where users agree to the license and terms of use. Users must apply separately for access to the NTU RGB+D 120 subset of the dataset through \href{https://rose1.ntu.edu.sg/dataset/actionRecognition/}{their webpage}.

\textbf{When will the dataset be distributed?} We are preparing the data and plan to release them at most upon publication.

\textbf{Will the dataset be distributed under a copyright or other intellectual property (IP) license, and/or under applicable terms of use (ToU)?}
% {If so, please describe this license and/or ToU, and provide a link or other access point to, or otherwise reproduce, any relevant licensing terms or ToU, as well as any fees associated with these restrictions.}
We will distribute the data with a CC BY-NC 4.0 license after filling a form where they agree to the license and terms of use. Users must apply separately for access to the NTU RGB+D 120 subset of the dataset.

\textbf{Have any third parties imposed IP-based or other restrictions on the data associated with the instances?}
% {If so, please describe these restrictions, and provide a link or other access point to, or otherwise reproduce, any relevant licensing terms, as well as any fees associated with these restrictions.}
The data from the NTU RGB+D 120 dataset have their own restrictions including redistribution, derivation or generation of a new dataset without permission and commercial usage.

\textbf{Do any export controls or other regulatory restrictions apply to the dataset or to individual instances?} No
%%%%%%%%%%%%%%%%%%%%%%%%%%%%%%%%%%%%%%%%%%

\subsection{Maintenance}

\textbf{Who will be supporting/hosting/maintaining the dataset?}
The authors and MARVL research group.

\textbf{How can the owner/curator/manager of the dataset be contacted (e.g., email address)?}
Through email (lmbravo@stanford.edu) and our GitHub repository.

\textbf{Is there an erratum?}
% {If so, please provide a link or other access point.}
No

\textbf{Will the dataset be updated (e.g., to correct labeling errors, add new instances, delete instances)?}{If so, please describe how often, by whom, and how updates will be communicated to users (e.g., mailing list, GitHub)?}
As the data is produced with the a data generation method all improvements in the method can lead to better products (e.g. posed meshes or contact annotations). We will communicate any updates to the data via the GitHub repository.

\textbf{If the dataset relates to people, are there applicable limits on the retention of the data associated with the instances (e.g., were individuals in question told that their data would be retained for a fixed period of time and then deleted)?}
% {If so, please describe these limits and explain how they will be enforced.}
No

\textbf{Will older versions of the dataset continue to be supported/hosted/maintained?}
% {If so, please describe how. If not, please describe how its obsolescence will be communicated to users.}
If updated will make versions of the dataset available but will maintain only the newest version.

\textbf{If others want to extend/augment/build on/contribute to the dataset, is there a mechanism for them to do so?}
% {If so, please provide a description. Will these contributions be validated/verified? If so, please describe how. If not, why not? Is there a process for communicating/distributing these contributions to other users? If so, please provide a description.}
Others are welcome to build on our work if abiding to our terms of use, license and use proper attribution. Any contributions can be discussed through our GitHub repository or by contacting the authors.

% \section{Dataset License}

% Add details of licensing. We are subject to the existing dataset's licenses, it looks like the only one with a specific one is NTU.
\section{LVLM Prompts}

\begin{figure}[ht]
    \centering
    \includegraphics[width=0.7\linewidth]{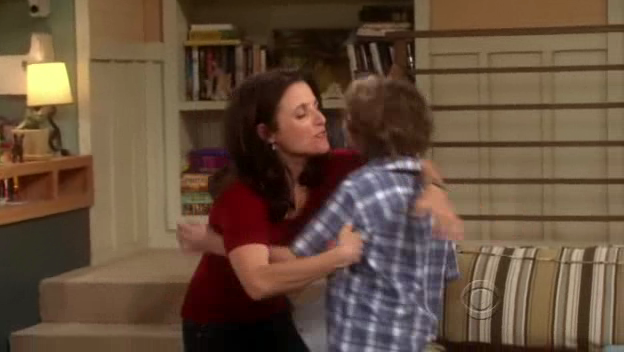}
    \caption{In-context example from the TV Interactions dataset provided with the prompt.}
    \label{fig:hug}
\end{figure}
To query the images in our dataset we use the prompt below.

\begin{tcolorbox}
\begin{verbatim}
Overview:
- You are participating in an image annotation project.
Your task is to annotate images where two people are interacting,
specifically identifying where their bodies touch.
Example:
- The first image is an example.
{"interaction": "hugging",
 "people": {"person_left": "woman wearing red hugging a child",
            "person_right": "child in a plaid shirt hugging a woman"},
 "orientation": "front to front",
 "contacts": [
    {"body_part_person_left":
        {"part_name: "upper arm", "body_side": "right"},
    "body_part_person_right":
        {"part_name: "forearm", "body_side": "left"},
    "confidence": 0.8},
    {"body_part_person_left":
        {"part_name: "hand", "body_side": "left"},
    "body_part_person_right":
        {"part_name: "back", "body_side": "right"},
    "confidence": 0.7}]
}             
Instructions:
1. Examine the second image carefully.
2. Annotate each point where body parts from the two individuals
make contact.
3. For each annotation, clearly specify:
    - Indicate which person (e.g., person on the left, person on the
    right) the body part belongs to.
    - The body part involved for each person and body side (either right
    or left or both)
    - The confidence level of that the contact is happening (0.0 - 1.0)
    
Output Requirements:
- Provide annotations in the following format:
{"interaction": "type of interaction",
 "people": {"person_left": "description of the person on the left",
           "person_right": "description of the person on the right"},
 "orientation": "orientation of the people (e.g., front to front,
 back to front, back to back, side to side)",
 "contacts": [
    {"body_part_person_left":
            {"part_name: "...", "body_side": "..."},
        "body_part_person_right":
            {"part_name: "...", "body_side": "..."},
        "confidence": 0.0 - 1.0},
    // More annotations here ]
}
- Use only this list of body part name: {body_parts}
Note:
- Aim for comprehensive coverage of all contact points, even those that
might appear minimal.
\end{verbatim}
\end{tcolorbox}

However, for the images that have interaction types we modify the prompt such that the instructions include the name of the interaction action (see box below).

\begin{tcolorbox}
\begin{verbatim}
Instructions:
1. Examine the second image of two people performing the action {action}
carefully.
2. Annotate each point where body parts from the two individuals
make contact.
3. For each annotation, clearly specify:
    - Indicate which person (e.g., person on the left, person on the
    right) the body part belongs to.
    - The body part involved for each person and body side (either right
    or left or both)
    - The confidence level of that the contact is happening (0.0 - 1.0)
\end{verbatim}
\end{tcolorbox}

For both cases we design the prompt such that the LVLM can make use of intermediate tasks to find the contacts. In particular, we: (1) provide an in-context example of two people hugging (Figure~\ref{fig:hug}) for which we detail the expected output; (2) request a description of the type of interaction and people involved in it. Even though we crop out the image to the minimum bounding containing both people of interest, there are instances with other subjects present. (3) The orientation of the people w.r.t one another, which in our experiments improves the labeling of the body parts in terms of chirality and recall.

% Show the effect of adding the action in the prompt for example images

% \begin{figure}[h]
%     \centering
%     % \includegraphics[width=\linewidth]{Styles//figures/method.png}
%     \includegraphics[width=\linewidth, height=10cm]{example-image-duck}
%     \caption{Examples of predicted contact map when adding action class to prompt}
%     \label{fig:}
% \end{figure}

\section{Additional qualitative results.}

\begin{figure}[h]
    \centering
    \includegraphics[width=\linewidth]{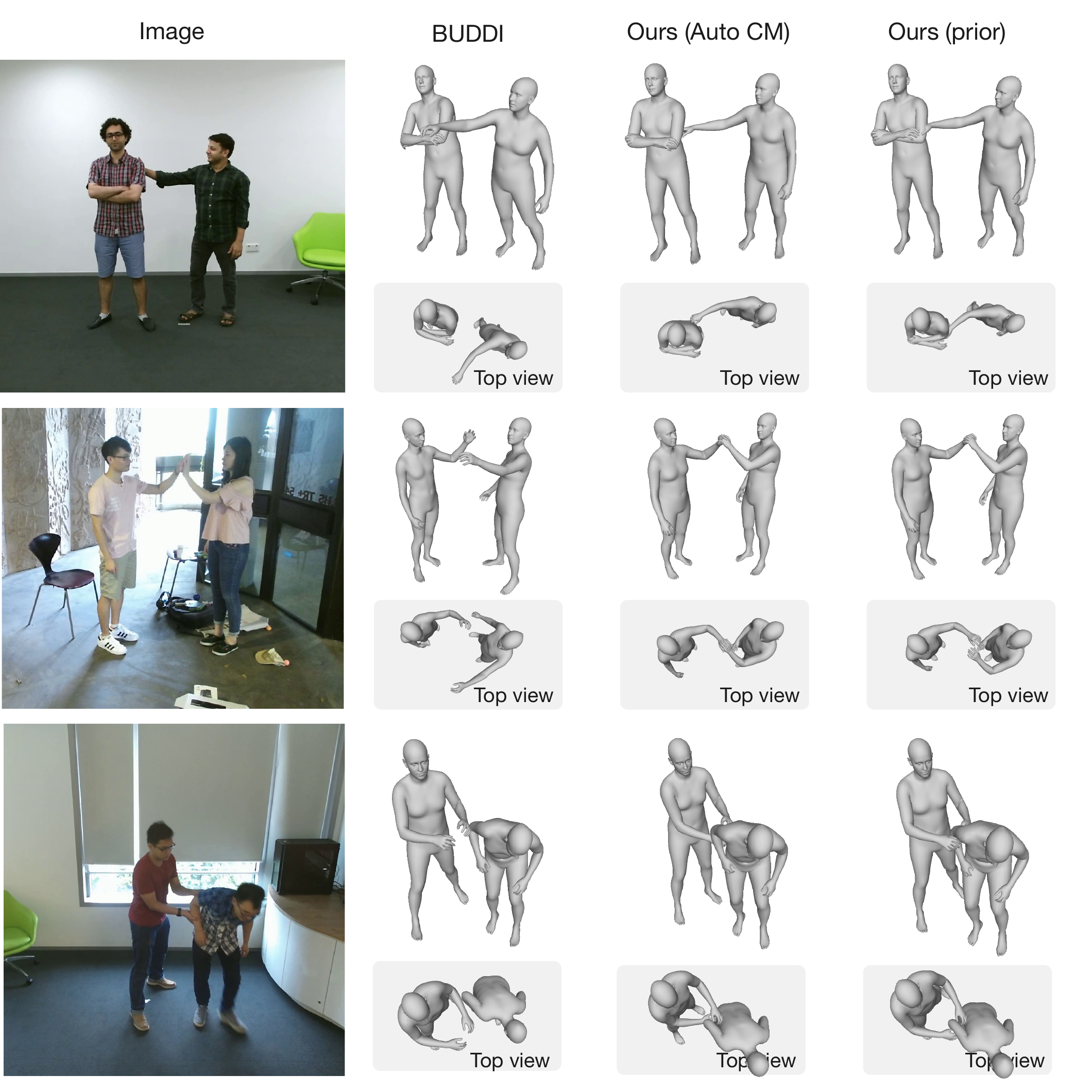}
    \caption{Examples of posed meshes on the close interactions NTU+RGBD 120 test set with BUDDI, Ours Auto CM, and Ours contact prior. Note the improved contact with our contact maps (Auto CM) and trained prior. \textit{Top row} pat on back. \textit{Middle row} high-five. \textit{Bottom row} Support someone.}
    \label{fig:ntu_qual_good}
\end{figure}

\begin{figure}[h]
    \centering
    \includegraphics[width=\linewidth]{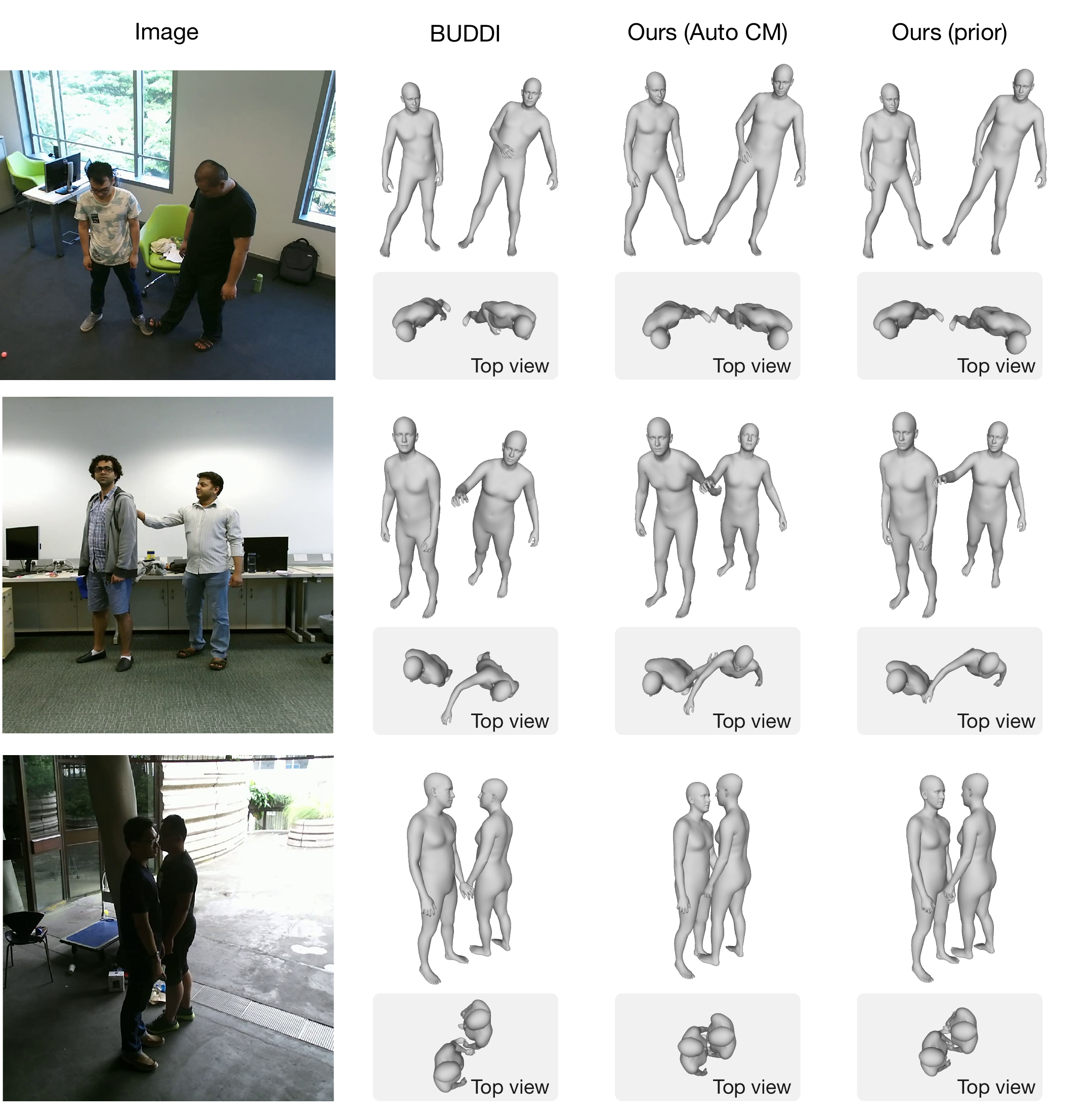}
    \caption{Examples of posed meshes on the close interactions NTU+RGBD 120 test set with BUDDI, Ours Auto CM, and Ours contact prior. \textit{Top row} the contact maps (Auto CM) can enforce contacts when the prior-based methods do not (Ours prior and BUDDI). \textit{Middle row} mistakes in laterality assignment of the contact maps lead to incorrect reconstructions, whereas the contact prior is robust to these cases. \textit{Bottom row} Occluded and dark scenes are challenging for all methods.}
    \label{fig:ntu_qual_bad}
\end{figure}

Figures \ref{fig:ntu_qual_good} and \ref{fig:ntu_qual_bad} show examples of posed meshes for images of the close interactions NTU+RGBD 120 test set. Overall the contact prior integrates the improved capacity of the contact maps (Auto CM) via training on the APU dataset in reflecting the image evidence and ensuring contact between surfaces. Mistakes in the contact maps can lead to incorrect reconstructions, which the trained prior is robust to.

\end{document}